\documentclass{article}
\usepackage{ijcai18}

\usepackage{url}
\usepackage{times}
\usepackage{xcolor}
\usepackage{soul}
\usepackage[utf8]{inputenc}
\usepackage[small]{caption}

\usepackage{amssymb}
\usepackage{xspace}
\usepackage{amsmath}
\usepackage{todonotes}
\usepackage{subfigure}
\usepackage{pgfplots}
\usepackage{array}

\newcolumntype{P}[1]{>{\centering\arraybackslash}p{#1}}

\usepackage{tikz}
\usetikzlibrary{shapes.geometric}
\usetikzlibrary{arrows,automata,shapes,calc,matrix,fit,arrows.meta}
\usetikzlibrary{positioning}

\usetikzlibrary{decorations.pathreplacing}
\makeatletter
\def\underbrace#1{%
	\@ifnextchar_{\tikz@@underbrace{#1}}{\tikz@@underbrace{#1}_{}}}
\def\tikz@@underbrace#1_#2{%
	\tikz[baseline=(a.base)] {\node[inner sep=4] (a) {\(#1\)};
		\draw[line cap=round,decorate,decoration={brace,amplitude=5pt}]
		(a.south east) -- node[below,inner sep=4pt] {\(\scriptstyle #2\)} (a.south west);}}

\title{Automated Verification of Neural Networks:\\ Advances, Challenges and Perspectives}

\author{
Francesco Leofante$^{1,4}$,
Nina Narodytska$^2$,
Luca Pulina$^3$,
Armando Tacchella$^1$
\\
$^1$ University of Genoa , $^2$ VMware Research\\
$^3$ University of Sassari,  $^4$ RWTH Aachen University\\
leofante@cs.rwth-aachen.de,
n.narodytska@vmware.com,
lpulina@uniss.it ,
armando.tacchella@unige.it
}

\newcommand{\ie}{\textit{i.e.}\xspace}
\newcommand{\eg}{\textit{e.g.}\xspace}

\definecolor{lightgreen}{rgb}{0.8,1.0,0.8}

\pgfplotsset{every axis/.append style={legend pos=north west,
		legend cell align={left},
		axis x line=middle,
		axis y line=middle,
		grid = major,
		width=16cm,
		height=8cm,
		grid style={dashed, gray!0},
		xmin=-4,     
		xmax= 4,    
		ymin=-1.5,     
		ymax= 2.5,   
		yticklabels={,,},
		xticklabels={,,},
		label style={font=\small},
		enlargelimits=false, scale=0.18 }}

\begin{document}

\maketitle

\begin{abstract}
Neural networks are one of the most investigated and widely used
techniques in Machine Learning. In spite of their success, they still
find limited application in safety- and security-related contexts,
wherein assurance about networks' performances must be provided. In
the recent past, automated reasoning techniques have been proposed
by several researchers to close the gap between neural networks and
applications requiring formal guarantees about their behavior. In this
work, we propose a primer of such techniques and a comprehensive
categorization of  existing approaches for the automated verification
of neural networks. A discussion about current limitations and 
directions for future investigation is provided to foster
research on this topic at the crossroads of Machine Learning
and Automated Reasoning.
\end{abstract}

\section{Introduction}
\label{sec:intro}

Neural Networks (NNs) are powerful learning models that can achieve
impressive results in many applications, such as image
classification~\cite{DBLP:conf/cvpr/TaigmanYRW14} or speech
recognition~\cite{DBLP:journals/taslp/YuHMCS12}, with some
architectures even claimed to be matching the cognitive abilities of 
humans~\cite{DBLP:journals/nature/LeCunBH15}.  
In spite of some exceptions --- see, \eg, \cite{jorgensen1997direct}
and more recently~\cite{DBLP:journals/corr/BojarskiTDFFGJM16,julian2016policy}
--- traditional applications of NNs have been mostly confined to
systems without safety or security requirements, due to the absence of
effective methods to guarantee the correct behavior of such models.  

There has long been an interest in the rigorous verification of NNs, 
with first attempts made in the early
2000s~\cite{938410,Pullum:2007:GVV:1199801}, mostly motivated by 
applications in avionic systems. This line of research was recently
refueled by critical discoveries made
in~\cite{DBLP:journals/corr/SzegedyZSBEGF13,DBLP:journals/corr/GoodfellowSS14}: 
machine learning models, including state-of-the-art Deep
Neural Networks (DNNs), can be unstable with respect
to \textit{adversarial perturbations}. Such perturbations represent
minimal changes to correctly classified input data that can cause a
network to respond in unexpected and incorrect ways. 
These discoveries confirmed the worthiness of efforts 
to develop techniques to provide guarantees about the
behavior of NNs and other learning models.

Among potential approaches to ensure correct behavior of NNs, those
based on Automated Reasoning show some promise. Since NNs are complex
implements, it is unlikely that their performances can be checked and
corrected manually. Techniques such as Adversarial
Training~\cite{DBLP:journals/corr/GoodfellowSS14} 
have been proposed with the intent to steer learning in the direction of
making resulting networks more robust to adversarial attacks. However,
recent results~\cite{DBLP:conf/ccs/Carlini017} have shown that existing
methods still lack thorough evaluations and often they are even unable
to detect adversarial examples. On the other hand, automated reasoning
tools can be applied 
to NNs ``out of the box'' to perform verification of desired
properties, \eg, robustness, safety, and equivalence. As with any
algorithmic technique, the challenge 
shifts towards the computational needs of automated verification, and the  
problem of scaling to networks of relevant size arises.   

%

Starting from the seminal contribution
of~\cite{DBLP:conf/cav/PulinaT10}, verification of NNs is not just a  
theoretical possibility, but it has witnessed diverse proposals based
on a variety of automated reasoning techniques, including
Boolean satisfiability (SAT) solvers, Satisfiability Modulo Theories
(SMT) solvers and Mixed Integer Programming (MIP)
solvers. The contributions to be found in the literature consider 
verification of diverse models, from conventional NNs, to
networks apt for representation learning, \ie, those
``...\emph{allowing a machine to be fed with raw data and automatically
discover the representations needed for detection or
classification}''~\cite{DBLP:journals/nature/LeCunBH15}.
Following the common usage found in the literature, we associate the
term \emph{deep} to networks apt for representation learning;
by contrast, we use the term \emph{shallow} to denote
networks designed within a conventional learning framework. As a
matter of fact, while all NNs are arranged in layers of elementary
computation units, conventional networks are indeed shallow 
since they rarely consist of several layers beyond input and output ones.
From the initial challenges and limitations presented
in~\cite{DBLP:journals/aicom/PulinaT12}, mostly related to the
application of SMT solvers to prove properties of shallow NNs,
several contributions have focused on the challenge of scaling SMT, as
well as SAT and MIP techniques to deep networks.
In this work, we present a survey of
such literature, and we contribute a categorization of existing
approaches based on properties and networks. To the best of our
knowledge, this is the first work attempting to put in perspective
this body of work within the communities of Machine Learning
and Automated Reasoning. Our contribution supports comparative  
assessments among applications, but it also helps in identifying open 
directions for future research. With the categorization herewith
proposed, we hope to lay the foundations for further innovation in
this interdisciplinary domain. 

The remainder of this paper is organized as follows. In
Section~\ref{sec:networks} we briefly introduce basic terms and
definitions about NNs; in Section~\ref{sec:decproc} we provide a short
introduction to the main automated reasoning techniques that have been
considered so far to verify NNs. Section~\ref{sec:stateofart} provides a
classification of the current relevant literature and
Section~\ref{sec:challenges} describes the current challenges and
provides some potential directions for future research.

\section{Feed-forward Neural Networks}
\label{sec:networks}
\begin{figure}[t]
	\begin{center}
		\scalebox{0.775}{ 
	\tikzstyle{line} = [-{Straight Barb[angle'=60,scale=0.5]}]
	\def\layersep{1.5cm}

	\begin{tikzpicture}[
	draw=black,
	node distance=\layersep,
	every pin edge/.style={{Straight Barb[angle'=60,scale=0.5]}- },
	neuron/.style={draw,circle,fill=black!0,minimum size=10pt,inner sep=0pt},
	input neuron/.style={neuron },
	output neuron/.style={neuron },
	hidden neuron/.style={neuron },
	annot/.style={text width=4em, text centered}
	]
	
	\foreach \name / \y in {1,...,3}
	\node[input neuron, pin=left:$x_\y$] (I-\name) at (0,-\y-0.5) {};
	
	\newcommand\Nhidden{3}
	
	\foreach \N in {1,...,\Nhidden} {
		\foreach \y in {1,...,5} {
			\path[yshift=0.5cm]
			node[hidden neuron] (H\N-\y) at (\N*\layersep,-\y cm) {};
		}
		\node[annot,above of=H\N-1, node distance=1cm] (hl\N) {Hidden layer \N};
	}
	
	\newcommand\Nout{2}
	
	\foreach \name / \y in {1,...,\Nout}
	\node[output neuron,pin={[pin edge={-{Straight Barb[angle'=60,scale=0.5]}}]right:$y_\y$}, right of=H\Nhidden-3] (O-\name) at (3*\layersep,-\y-1) {};
	
	\foreach \source in {1,...,3}
	\foreach \dest in {1,...,5}
	\path[line] (I-\source) edge (H1-\dest);
	
	\foreach [remember=\N as \lastN (initially 1)] \N in {2,...,\Nhidden}
	\foreach \source in {1,...,5}
	\foreach \dest in {1,...,5}
	\path[line] (H\lastN-\source) edge (H\N-\dest);
	
	\foreach \source in {1,...,5}
	\foreach \dest in {1,...,\Nout}
	\path[line] (H\Nhidden-\source) edge (O-\dest);
	
	
	\node[annot,left of=hl1] {Input layer};
	\node[annot,right of=hl\Nhidden] {Output layer};
	\end{tikzpicture}
 }
	\end{center}
	\vspace*{-2ex}
	\caption{A simple network containing 3 input nodes, 2 output nodes and 3 hidden layers containing 5 nodes each.}
	\label{fig:nnet}
\end{figure}
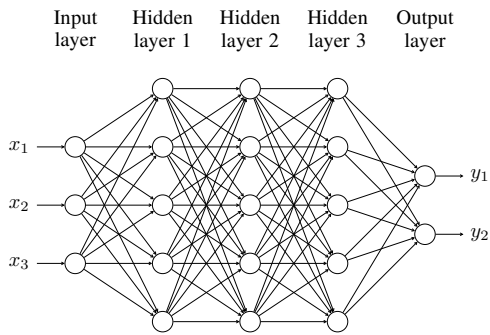

This survey will focus on approaches developed for the verification of
feed-forward NNs. In such networks, \textit{neurons} are arranged in
disjoint \textit{layers}, with each layer being fully connected with
the next one, but without connection between neurons in the same
layer. We call a layer without incoming 
connections \textit{input} layer, a layer without outgoing connections
\textit{output} layer, while all other layers are referred to as
\textit{hidden} layers. Connections between neurons in the network are
labeled with \textit{weights} which can take, in the most general
case, real values\footnote{We do not discuss here how NNs are
  learned. Frameworks such as \texttt{Keras}~\cite{chollet2015keras}
  provide ready-to-use solutions for learning networks from
  data.}. Furthermore, each neuron is characterized by an
\textit{activation function} defining the input-output relation for
that particular neuron. A pictorial representation of the architecture
described can be see in Fig.~\ref{fig:nnet}. 

At a high level, these networks can be seen as functions 
$\nu : I^n \to O^m$, mapping an $n$-dimensional \emph{input domain}
$I^n$ ($n > 0$) to a $m$-dimensional \emph{output domain} $O^m$ ($m
>0$). We argue that this representation captures most cases of practical
interest. For instance, a network computing an approximation
of some function $f: \mathbb{R}^n \to \mathbb{R}$ would have $I = O =
\mathbb{R}$, whereas a network classifying 8-bit images of size $h \times v$ in
 two classes would be defined as ${\nu: \{0,\ldots,255\}^{h \cdot v}
\to \{0, 1\}}$ with $I=\{0, \ldots, 255\}$ and $O = \{0,1\}$. 
The mapping is performed by feeding an input $e \in I^n$ to the
network through its input layer, which is then propagated to the
output layer by successively computing linear combinations of values
from nodes  in the preceding layer and applying activation functions
to the result. Several types of activations exist, each of them having
different properties which determine the expressive power of the
network. Most commonly used activation functions include, \eg,
logistic sigmoid, hyperbolic tangent and rectified linear units (ReLU)
-- see Fig.~\ref{fig:activations}. 

\begin{figure}[t]
	\centering
	\scalebox{0.9}{
	{\def\arraystretch{2.5}
	\tikzstyle{line} = [-{Straight Barb[angle'=60]}]
	\tikzset{font={\fontsize{30pt}{12}\selectfont}}
	\begin{tabular}{| P{2.6cm} | P{2.6cm} | P{2.6cm} |}
		\hline
		\textbf{Logistic} & \textbf{Hyperbolic tangent} & \textbf{ReLU}\\
		\hline
		
		& & \\[-15pt]
		
		\scalebox{0.3}{
		\begin{tikzpicture}[thick]
		\node (a) {$x_1$};
		\node[below of=a] (b) {$\vdots$};
		\node[below of=b] (c) {$x_i$};
		\node[below of=c] (d) {$\vdots$};
		\node[below of=d] (e) {$x_n$};
		
		\node[right of=c, draw, circle, node distance=4cm ] (neuron) {
			
			\begin{tikzpicture}
			\begin{axis}
			\addplot[ultra  thick,samples=500] {1/(1+exp(-x))};
			\end{axis}
			\end{tikzpicture}
		};
		
		\node[right of=neuron, node distance=3cm] (output) {$y$};
		
		\path[line,shorten <=0.5cm] (a) edge (neuron);
		\path[line,shorten <=0.5cm] (b) edge (neuron);
		\path[line,shorten <=0.3cm] (c) edge (neuron);
		\path[line,shorten <=0.5cm] (d) edge (neuron);
		\path[line,shorten <=0.5cm] (e) edge (neuron);
		\path[line] (neuron) edge (output);
	\end{tikzpicture}}	 &
		
		\scalebox{0.3}{
			\begin{tikzpicture}[thick]
			
			\node (a) {$x_1$};
			\node[below of=a] (b) {$\vdots$};
			\node[below of=b] (c) {$x_i$};
			\node[below of=c] (d) {$\vdots$};
			\node[below of=d] (e) {$x_n$};
			
			\node[right of=c, draw, circle, node distance=4cm] (neuron) {

				\begin{tikzpicture}
				\begin{axis}
				\addplot[ultra  thick,samples=500] {tanh(x)};
				\end{axis}
				\end{tikzpicture}
			};
			
			\node[right of=neuron, node distance=3cm] (output) {$y$};
			
			\path[line,shorten <=0.5cm] (a) edge (neuron);
			\path[line,shorten <=0.5cm] (b) edge (neuron);
			\path[line,shorten <=0.3cm] (c) edge (neuron);
			\path[line,shorten <=0.5cm] (d) edge (neuron);
			\path[line,shorten <=0.5cm] (e) edge (neuron);
			\path[line] (neuron) edge (output);
		\end{tikzpicture}} &
		
		\scalebox{0.3}{
			\begin{tikzpicture}[thick]
			
			\node (a) {$x_1$};
			\node[below of=a] (b) {$\vdots$};
			\node[below of=b] (c) {$x_i$};
			\node[below of=c] (d) {$\vdots$};
			\node[below of=d] (e) {$x_n$};
			
			\node[right of=c, draw, circle, node distance=4cm] (neuron) {

				\begin{tikzpicture}
				\begin{axis}
				\addplot[ultra  thick,samples=500] {max(0, x)};
				\end{axis}
				\end{tikzpicture}
			};
			
			\node[right of=neuron, node distance=3cm] (output) {$y$};
			
			\path[line,shorten <=0.5cm] (a) edge (neuron);
			\path[line,shorten <=0.5cm] (b) edge (neuron);
			\path[line,shorten <=0.3cm] (c) edge (neuron);
			\path[line,shorten <=0.5cm] (d) edge (neuron);
			\path[line,shorten <=0.5cm] (e) edge (neuron);
			\path[line] (neuron) edge (output);
		\end{tikzpicture}}\\
		\hline
		 \footnotesize \(\displaystyle y = \frac{1}{1+e^{-\Sigma_i \text{ } x_i}} \) &\footnotesize $ y = tanh\left(\Sigma_i \text{ } x_i\right) $ &\footnotesize $ y = max\left(0,\Sigma_i \text{ } x_i\right) $\\[4pt]
		 \hline	
	\end{tabular}}}
	\caption{Examples of commonly used activation functions, assuming that input weights are fixed to one.}
	\label{fig:activations}
\end{figure}

\section{Decision procedures}
\label{sec:decproc}
Several approaches have been proposed to verify different classes of networks. Even though such approaches might differ in several aspects, they all tackle the verification problem by encoding networks to constraint systems. In this section
we briefly introduce the decision procedures that have been most commonly used to solve constraint systems encoding neural networks. For further details we refer the interested reader to~\cite{DBLP:series/faia/2009-185} and~\cite{Schrijver}.

\begin{figure}[b]
	\begin{center}
		\scalebox{0.9}{\begin{tikzpicture}

	\node (formula) {Input CNF formula} ;
	
	\node[draw, rounded corners, below of=formula, node distance=1.18cm] (bcp) {BCP};
	
	\node[draw, ellipse, below of=bcp, node distance=1.7cm] (conflict) {Conflict?}; 
	
	\node[draw, rounded corners, right of=conflict, node distance=2.5cm](resolution) {Resolution};
	
	\node[draw, ellipse, above of=resolution, node distance=1.7cm] (resolved) {Resolved?};

	\node[draw, ellipse, below of=conflict, aspect=2, node distance=1.8cm ] (allass) { All assigned?}; 
	
	\node[draw, rounded corners, left of=allass, node distance=2.8cm](decision) {Decision};
	
	\node[draw, rounded corners, right of=allass, node distance=5cm] (sat) {SAT};
	
		\node[draw, rounded corners, above of=sat, yshift=2.5cm](unsat) {UNSAT};
	
	
		\path [draw, -latex'] (formula) -- (bcp);
		
		\path [draw, -latex'] (bcp) -- (conflict);
		
		\path [draw, -latex'] (conflict) -- node [near start, above] {y} (resolution);
		
		\path [draw, -latex'] (conflict) -- node [near start, right] {n} (allass);
		
		\path [draw, -latex'] (resolution) -- (resolved);
		
		\path [draw, -latex'] (resolved) -- node [near start,above] {n} (unsat);
		
		\path [draw, -latex'] (allass) -- node [near start, above] {y} (sat);
		
		\path [draw, -latex'] (allass) -- node [near start, above] {n} (decision);
		
		\path [draw, -latex'] (decision) |- (bcp);
		
		\path [draw, -latex'] (resolved) -- node [near start,above] {y} (bcp);

\end{tikzpicture}}
	\end{center}
	\vspace*{-2ex}
	\caption{The CDCL framework.}
	\label{fig:dpll}
\end{figure}
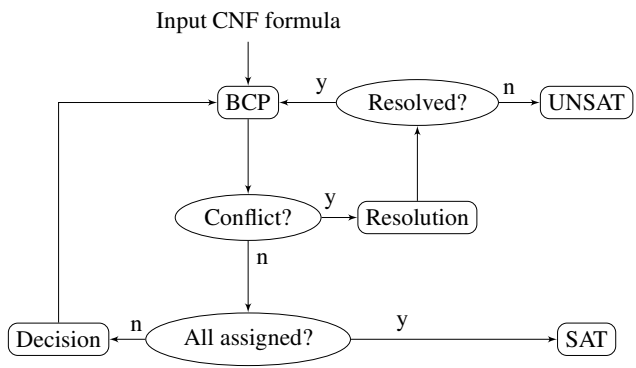

\paragraph{SAT} SAT solving aims to check the satisfiability of a
propositional logic formula $\varphi$ represented as Boolean combinations of atomic (Boolean)
propositions. Although several algorithms have been proposed to solve the boolean
satisfiability problem, here we introduce CDCL-style SAT solving algorithm (see Fig.~\ref{fig:dpll}), being the most commonly
implemented in state-of-the-art SAT solvers.

The CDCL algorithm proceeds as follows. Starting from an input CNF formula, the algorithm explores the search space by iteratively making decisions, \ie, it assigns truth values to some heuristically chosen propositions.
 After each such decision, the algorithm applies Boolean Constraint Propagation
(BCP) to determine further variable assignments that are implied by the last
decision. If BCP leads to a conflict, \ie, if the value of a proposition is implied to
be true as well as false at the same time, \textit{conflict-driven clause-learning} and
\textit{non-chronological backtracking} are applied: the algorithm follows back
the chain of implications and applies resolution to derive a reason for
the conflict in form of a conflict clause, which is added to the solver’s clause
set. Backtracking removes previous decisions and their implications until the
conflict clause can be satisfied.

If the input has clauses consisting of a single literal, these literals will be directly
assigned. Therefore, the algorithm starts with BCP to detect
implications. If BCP leads to a conflict, the algorithm tries to resolve the conflict.
If the conflict cannot be resolved, the input formula is unsatisfiable. Otherwise,
if the conflict is successfully resolved, the algorithm backtracks and continues
with BCP. If BCP can be completed without any conflicts, a new decision
is made if there are any unassigned propositions. Otherwise, a satisfying
solution is found.

\paragraph{SMT} Satisfiability Modulo Theories is the problem of deciding the
satisfiability of a first-order formula with respect to some decidable
theory $\mathcal{T}$. In particular, SMT generalizes the boolean
satisfiability problem (SAT) by adding background theories such as the
theory of real numbers, the theory of integers, and the theories of
data structures (\textit{e.g.}, lists, arrays and bit vectors).

\begin{figure}[t]
	\begin{center}
	\scalebox{0.9}{\begin{tikzpicture}[thick]

\pgfdeclarelayer{background}
\pgfdeclarelayer{foreground}
\pgfsetlayers{background,main,foreground}

\tikzstyle{doc}=[%
draw,
thick,
align=center,
color=black,
shape=document,
minimum width=10mm,
minimum height=15.2mm,
shape=document,
inner sep=2ex,
]

    \node [rectangle, draw,text centered, rounded corners, text width=10em, minimum height=5em, minimum width = 10em] (sat) {    	
    	\begin{tabular}{c}
    	Boolean abstraction\\
    	\\
    	SAT solver
    	\end{tabular}};

    \node[above of=sat, node distance=1.8cm](cnf){\begin{tabular}{c}
    	Input CNF\\[-0.08cm]
    	formula
    	\end{tabular}};
    
    \begin{pgfonlayer}{foreground}
    \path (sat.west |- sat.west) node (a) {};
    \path (sat.east -| sat.east)node (b) {};
    \path[ draw=black!50, dashed]
    (a) rectangle (b);
    \end{pgfonlayer}
    
    \node [rectangle, draw,text centered, rounded corners, thick, text width=6.5em, minimum height=5em, minimum width =6.5em, right of=sat, node distance=4cm] (smt) {Theory solver};

	\node[below left of=sat, node distance=2.6cm](res){\begin{tabular}{c}
		SAT \\[-0.2cm]
		or\\[-0.1cm]
		UNSAT
		\end{tabular}};

	\path [draw, -latex'] (cnf) -- (sat);
	\draw[-latex,bend left=45]  (sat) edge [above] node {constraints} (smt);
	\draw[-latex,bend right=-45]  (smt) edge  [below] node {\begin{tabular}{c}
		SAT + model \\
		UNSAT + explanation
		\end{tabular}} (sat);
	\path [draw, -latex'] (sat) -- (res);
	
\end{tikzpicture}}
	\end{center}
	\vspace*{-2ex}
	\caption{The SMT solving framework.}
	\label{fig:smt}
\end{figure}
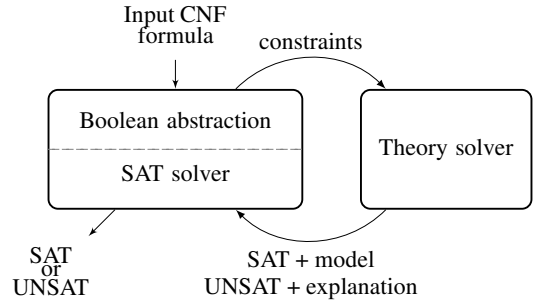

To decide the satisfiability of an input formula $\varphi$ in CNF, SMT solvers (see Fig. \ref{fig:smt}) typically first build a \emph{Boolean abstraction} $\textit{abs}(\varphi)$ of $\varphi$ by replacing each constraint by a fresh Boolean variable (proposition), \eg,

\begin{eqnarray*}
\arraycolsep=2pt
\begin{array}{ccccccccccc}
\varphi &: &\underbrace{x \geq y} &\wedge &(&\underbrace{y > 0}& \vee &\underbrace{x >0}&)& \wedge &\underbrace{y \leq 0} \\
\textit{abs}(\varphi)&:&A& \wedge& (&B& \vee &C&)& \wedge  &\neg B
\end{array}
\label{eq:abs}
\end{eqnarray*}
where $x$ and $y$ are real-valued variables, and $A$, $B$ and $C$ are propositions.

\noindent A SAT solver searches for a
satisfying assignment $S$ for $\textit{abs}(\varphi)$, \eg, $S(A)=1$, $S(B)=0$,
$S(C)=1$ for the above example.  If no such assignment exists then the
input formula $\varphi$ is unsatisfiable. Otherwise, the
consistency of the assignment in the underlying theory
is checked by a \emph{theory solver}. In our example, we check whether the set $\{ x \geq y,\ y
\leq 0,\ x > 0\}$ of linear inequalities is feasible, which is the
case. If the constraints are consistent then a satisfying solution
(\textit{model}) is found for $\varphi$. Otherwise, the theory solver returns a theory lemma $\varphi_E$ giving an
\textit{explanation} for the conflict, \eg, the negated conjunction some inconsistent input constraints.
The explanation is used to refine the Boolean abstraction $\textit{abs}(\varphi)$ to $\textit{abs}(\varphi)\wedge \textit{abs}(\varphi_E)$.
These steps are iteratively executed until either a theory-consistent
Boolean assignment is found, or no more Boolean satisfying assignments
exist.

\paragraph{MIP}
Mixed Integer Linear Programming (MIP) solves linear problems over a set of integer and real valued variables.
MIP contains a set of \textit{decision variables}, a set of \textit{linear constraints} over these variables and
an \textit{objective function} to be optimized (minimized or maximized) that is linear in decision variables.
Without loss of generality we consider a minimization formulation of a MIP and assume that all variables are integers that take values in a given interval.
Let $x_1,\ldots, x_n$ be a set of decision variables, an integer linear program can be written as
\begin{align}
\text{min} & \sum_{i=1} c_ix_i \nonumber \\
\text{subject to } & \sum_{i=1} a_{ji}x_i \geq b_j, j \in [1,m]  \nonumber \\
& x_i \in [a_i,b_i]\cap \mathbb{Z}, i \in [1,n]   \nonumber
\label{eq:mip}
\end{align}

Values $c_i,a_{ij}$ and $b_j$ are constants that are  specified during  problem formulation.
One general approach to solving MIPs is by using the \emph{branch-and-cut} method that employs \emph{branch-and-bound} and \emph{cutting planes} techniques.
The branch-and-bound method performs two main steps. First, it solves a linear relaxation of MIP where all integrality constraints are relaxed. In other words, we assume that all integer variables can take real values. The cost of a solution of the relaxed problem gives a lower bound on the optimal solution of the original problem. However, this solution can contain fractional values.
Therefore, the MIP solver has to branch on one of variables with the fractional value, splitting the search space into two parts.
For example, if $x_2 = 0.5$ in a solution then the split is $x_2=0$ or $x_2=1$.
Based on these decision points, the search procedure builds a branching tree and stores the best solution found in each node.
These solutions are used to prune future branches. The cutting planes technique is used to cut off fractional solutions. These cutting inequalities are learned during search and help to improve the quality of Linear Programming relaxations.

\section{State of the art: a bird's eye view}
\label{sec:stateofart}
\begin{table*}
  \scriptsize
  \begin{center}
  \begin{tabular}{|l|c|c|c|c|}
    \hline
    \textbf{Network} &
    \multicolumn{2}{c|}{\textbf{Invariance}} &
    \textbf{Invertibility} &
    \textbf{Equivalence} \\ \cline{2-3}
    &
    \textbf{Local} & \textbf{Global} & & \\ \hline
    BNN &
    \parbox[c][10ex][c]{0.18\textwidth}{\cite{DBLP:journals/corr/abs-1709-06662}} &
    \parbox[c][10ex][c]{0.18\textwidth}{\cite{DBLP:journals/corr/abs-1710-03107}} &
    \parbox[c][10ex][c]{0.18\textwidth}{\cite{DBLP:journals/corr/abs-1802-08795}} &
    \parbox[c][10ex][c]{0.18\textwidth}{\cite{DBLP:journals/corr/abs-1709-06662}}
    \\ \hline
    DNN(ReLU) &
    \parbox[c][10ex][c]{0.18\textwidth}{\cite{DBLP:journals/corr/abs-1710-00486,DBLP:conf/cav/KatzBDJK17,DBLP:conf/nips/BastaniILVNC16}} &
    \parbox[c][10ex][c]{0.18\textwidth}{\cite{DBLP:conf/cav/KatzBDJK17,DBLP:journals/corr/abs-1709-09130,DBLP:journals/corr/abs-1801-05950}} &
    \parbox[c][10ex][c]{0.18\textwidth}{---} &
    \parbox[c][10ex][c]{0.18\textwidth}{---}
    \\ \hline
    DNN(ReLU+Pooling) &
    \parbox[c][10ex][c]{0.18\textwidth}{\cite{DBLP:conf/atva/Ehlers17,DBLP:journals/corr/abs-1711-07356}} &
    \parbox[c][10ex][c]{0.18\textwidth}{\cite{DBLP:journals/corr/abs-1711-00455,DBLP:journals/corr/abs-1712-06174}} &
    \parbox[c][10ex][c]{0.18\textwidth}{\cite{DBLP:conf/atva/Ehlers17}} &
    \parbox[c][10ex][c]{0.18\textwidth}{---}
    \\ \hline
    DNN &
    \parbox[c][10ex][c]{0.18\textwidth}{\cite{DBLP:conf/cav/HuangKWW17}} &
    \parbox[c][10ex][c]{0.18\textwidth}{---} &
    \parbox[c][10ex][c]{0.18\textwidth}{---} &
    \parbox[c][10ex][c]{0.18\textwidth}{---}
    \\ \hline
    NN &
    \parbox[c][10ex][c]{0.18\textwidth}{\cite{DBLP:journals/aicom/PulinaT12,cheng2017maximum}} &
    \parbox[c][10ex][c]{0.18\textwidth}{\cite{DBLP:conf/cav/PulinaT10,scheibler2015towards}} &
    \parbox[c][10ex][c]{0.18\textwidth}{---} &
    \parbox[c][10ex][c]{0.18\textwidth}{---}
    \\ \hline
  \end{tabular}
  \end{center}
  \caption{\label{tab:literature}Literature classified by network type
    (rows) and properties considered (columns). For invariance
    properties, a further distinction is made between papers dealing
    with local or global invariance. ``BNN'' row collects references
    dealing specifically with Binarized (deep) NNs. Deep network
    references are organized according to the kind of nodes for which
    the techniques thereto proposed are applicable: ``DNN(ReLU)'' for networks
    made up of ReLUs only, ``DNN(ReLU+Pooling)'' if also pooling
    nodes are considered, and ``DNN'' if no restriction is placed on nodes.
    ``NN ``row collects references which, in principle,
    consider any kind of NN, but are
    mostly concerned with shallow networks.}
\end{table*}

To describe properties of NNs defined as $\nu : I^n \to O^m$, 
let $pre(x)$ and $post(y)$ be sorted first order logic formulas, with
$x$ and $y$ occurring as free variables of sort $S_I$ and $S_O$,
respectively. Informally, $S_I$ is the input type and
$S_O$ is the output type required by the network thought as a function
in some programming language.  
We say that $pre$ defines \emph{preconditions} on the
input of a network, and $post$ defines \emph{postconditions} on  its
output. We consider interpretations that map sort $S_I$ to the input
domain $I^n$ and sort $S_O$ to the output domain $O^m$. For instance,
an interpretation in a network defined as $\nu: \mathbb{R} \to \{0,1\}$,
could map $S_I$ = \emph{real} and $S_O$ = \emph{boolean} to
corresponding domains $\mathbb{R}$ and $\{0, 1\}$. 
Interpretations map variables $x$ and $y$ to values  in the domains
$I^n$ and $O^m$. We write ${\cal I}(x \to e)$ to denote that variable
$x$ is mapped to value $e \in I^n$ by interpretation ${\cal I}$, and
$\varphi^{\cal I}$ to denote the value of expression $\varphi$ under
interpretation $\cal  I$. We also consider the predicates
``$=$'', ``$\neq$'', and ``$<$'' with the usual semantics.

\paragraph{Properties.} To the extent of our knowledge, all the
studies published so far about 
automated verification of NNs, focused on three kinds of
properties:
\begin{itemize}
\item \emph{Invariance}.
  For specific conditions $pre$ and $post$, asserting an invariance
  property for a network $\nu$ amounts to state
  \begin{equation}
    \label{eq:invar}
    \forall x. \forall y. (pre(x) \wedge y =  \nu(x)) \implies post(y)
  \end{equation}
  The goal of automated verification is to prove (\ref{eq:invar}) or
  find a \emph{counterexample}, \ie, some value $e \in I^n$ such that
  $(pre(x) \wedge \neg post(y)^{{\cal I}(x \to e, y \to \nu(e))}$ is true.

\item \emph{Invertibility}. For specific conditions $pre$ and $post$,
  asserting an invertibility property for a network $\nu$ amounts to
  state
  \begin{equation}
    \label{eq:invert}
    \forall y. \exists x. (post(y) \wedge y = \nu(x)) \implies pre(x)
  \end{equation}
  Proving invertibility might be less interesting than actually
  finding a specific \emph{realization}, \ie, given an output pattern
  $p \in O^m$ find an input pattern $e \in I^n$ such that
  $(post(y)  \wedge y=\nu(x) \wedge pre(x))^{{\cal I}(x \to e, y \to p)}$ is true.

\item \emph{Equivalence}. While invariance and invertibility refer to a
  single network, equivalence is a property involving two networks
  $\nu$ and $\nu'$. For specific conditions $pre$ and $post$, it is
  defined as
  \begin{equation}
    \label{eq:equiv}
    \begin{array}{r@{\,}l}
      \forall x. \forall y. \forall w( & pre(x) \wedge \\
      & y = \nu(x) \wedge post(y) \wedge \\
      & w = \nu'(x) \wedge post(w)) \implies y = w
    \end{array}
  \end{equation}
  The property can be either proved as such or a \emph{counterexample}
  can be produced, \ie, some $e \in I^n$ such that
  $$
  \begin{array}{r@{}l}
    ( & pre(x) \wedge \\
    & post(y) \wedge post(w) \wedge
    y \neq w)^{{\cal I}(x \to e, y \to \nu(e), w \to \nu'(e))}
    \end{array}
  $$
  is true. In contexts wherein strict equality might be inappropriate,
  we can replace the term $y = w$ in (\ref{eq:equiv}) with the term
  $|| y - w || < \epsilon$, assuming that $|| \cdot ||^{\cal I}$ is
  a norm over $O^m$ and $\epsilon^{\cal I} \in O^m$ is a 
  a \emph{tolerance}, \ie, a threshold under which the
  response of the networks is considered to be indistinguishable. 
\end{itemize}
As stated in (\ref{eq:invar}) to (\ref{eq:equiv}), the scope of the
properties is \emph{global}, \ie, interpretations range over full
domains $I^n$ and $O^m$. Researchers have also considered \emph{local}
versions wherein interpretations range over specific regions of the
input and output domains.

\paragraph{Literature.}In Table~\ref{tab:literature} we have organized
all the contributions 
found in the literature, wherein the techniques described in
Section~\ref{sec:decproc} are utilized to prove properties about
NNs. For invariance, we have considered an additional
classification into global and local versions (first two columns of
the Table), whereas for invertibility and equivalence, we did not make
a distinction due to the limited number of references available. In
particular,~\cite{DBLP:conf/atva/Ehlers17} mentions a local flavor
of invertibility considering DNNs with ReLU and
MaxPool nodes, whereas~\cite{DBLP:journals/corr/abs-1802-08795}
considers a global flavor of the same property in the context of
binarized (deep) NNs (BNNs). As for equivalence, we found only
one contribution about BNNs
in~\cite{DBLP:journals/corr/abs-1709-06662}.

The first paper to consider automated verification for
shallow networks is~\cite{DBLP:conf/cav/PulinaT10}. Indeed,
some authors refer to~\cite{938410} as the first attempt to scrutinize
NNs in order to give formal guarantees about their
performances. We must stress here that while~\cite{938410} put forth
the first analytical approach to verify network's accuracy,
it did not consider algorithmic verification which, on the
contrary, is the focus of~\cite{DBLP:conf/cav/PulinaT10} and
the ensuing literature herewith considered. In
particular,~\cite{DBLP:conf/cav/PulinaT10} focused on a global
invariance condition for multi-input, single-output networks involving non-linear activation functions, whereby
given $\nu: I^n \to O$, as long as the input $e \in I^n$ is guaranteed
to range within some prescribed interval, then $\nu(e) \in [a,b]$ with
$a,b \in O$. 
Also for the first
time,~\cite{DBLP:conf/cav/PulinaT10} deals with verification-triggered network \emph{repair},
\ie, how to modify network's weights in order to meet the invariance
condition. The initial contribution is extended
in~\cite{DBLP:journals/aicom/PulinaT12} to consider other invariance
conditions, including local invariance and global sensitivity. The
main limitation of these early attempts is that the number of
activations functions in the networks is relatively small (in the
order of tens), and the proposed encoding hardly scales for the kind
of (deep) networks considered nowadays in applications. Other attempts
at proving invariance properties of shallow NNs
include~\cite{scheibler2015towards} and~\cite{cheng2017maximum}. The
former approach leverages SMT technology over non-linear arithmetic
--- with essentially the same limitations exposed
in~\cite{DBLP:journals/aicom/PulinaT12} --- whereas the latter
considers MIP techniques.

More recently, due to the growing interest in DNNs,
several papers came out proposing approaches that, to some extent, can
deal with networks having nodes in the order of thousands, \eg,
1800 ReLU nodes in~\cite{DBLP:conf/cav/KatzBDJK17}, and millions of
parameters, \eg, about 1.25 million parameters in the CIFAR
experiment considered in~\cite{DBLP:conf/cav/HuangKWW17}. Most of the 
contributions proposed in the literature
focus on invariance properties only and consider
restrictions of DNNs, \ie, they consider only some
kind of activation function. ReLUs, possibly with MaxPool nodes, are by
far the most common target: to the best of our knowledge, the only
contribution dealing with convolutional, ReLU, max-pooling, 
and softmax layers is~\cite{DBLP:conf/cav/HuangKWW17}, whose goal
is to verify (local) adversarial robustness using SMT technology black-box.
Noticeably,~\cite{DBLP:journals/corr/abs-1711-07356,DBLP:conf/nips/BastaniILVNC16}
and~\cite{DBLP:journals/corr/abs-1712-06174} are the only
contributions using MIP encodings to prove adversarial robustness in a
local and global fashion, respectively. Other contributions instead consider variations of SMT technology wherein the
theory solver is specialized to deal with the problem at hand. This is
the case
of~\cite{DBLP:journals/corr/abs-1710-00486,DBLP:conf/cav/KatzBDJK17,DBLP:journals/corr/abs-1801-05950}
which consider Reluplex, a Simplex-based decision procedure specialized to tackle
constraints arising from the verification of deep networks comprised of
ReLU nodes. Uniquely among other surveyed
works,~\cite{DBLP:journals/corr/abs-1710-00486} proposed a combination
between inductive and deductive techniques to improve scaling on large
networks: a global invariance check is reduced to a series of local
invariance check using clustering techniques on the input
space. In~\cite{DBLP:conf/atva/Ehlers17} networks made of ReLUs and
MaxPool nodes are considered. Also in this case, an original
combination of SAT and a theory solver (an Integer Linear Programming
engine) is considered to enable verification of a network containing
1341 nodes.

Overall, looking at Table~\ref{tab:literature}, it is clear that
existing literature leaves a lot of potential areas of interest to be
covered. Firstly, full-fledged DNNs are considered
only in one contribution, namely~\cite{DBLP:conf/cav/HuangKWW17}, and
the approach therewith proposed is still in its prototypical
stage. This is in stark contrast with the development of tools for
\emph{learning} neural networks, many of which are available
off-the-shelf, reaching a substantial degree of
sophistication. Secondly, network equivalence and invertibility
remain mostly uncharted, but they could support effective application
of neural networks in many ways. For instance, showing that a
small-footprint network is equivalent to a large-footprint one, memory
and energy could be saved by running the smallest network. Also
showing that a given output pattern may not be produced by any input
pattern, could help debug networks that could otherwise reach
production based solely on empirical tests.

\section{Challenges and perspectives}
\label{sec:challenges}
The first and foremost challenge in automated verification of NNs, is
to coordinate the efforts of two ``separated at birth'' AI
communities: Machine Learning and Automated Reasoning.
On one side, Machine Learning has made remarkable progresses in the
last decade, and it is now one of the mainstream AI domains, backed by
substantial funding and success stories, \eg, Deepmind's
AlphaGo\footnote{\scriptsize
  \url{https://deepmind.com/blog/alphagos-next-move/}.},
Facebook's DeepFace~\cite{DBLP:conf/cvpr/TaigmanYRW14}, and
Neurala's\footnote{\scriptsize \url{https://www.neurala.com/press-releases/edge-deep-learning-without-cloud}.} Lifelong-DNN\textsuperscript{TM}, to cite only some.
The existing need for explanation and certification associated with
learning algorithms has been recognized by the community, \eg, by a
number of recent workshops dedicated to the
topic~\footnote{\scriptsize Explainable Artificial Intelligence at IJCAI 2017,
  \url{http://home.earthlink.net/~dwaha/research/meetings/ijcai17-xai/}
  and Interpretable Machine Learning at NIPS 2017,
  \url{http://interpretable.ml/}} and DARPA research
programs~\footnote{\scriptsize Explainable Artificial Intelligence
  (XAI)\url{https://www.darpa.mil/program/explainable-artificial-intelligence}
}.
For NNs, most researchers in Machine Learning agree that new
techniques are needed to understand, trust and manage networks that
might be used in regulated areas, such as law or medicine, or in
safety- and security-critical applications.  As noted by Percy Liang
during his invited talk at AAAI 2018\footnote{\scriptsize "How Should We Evaluate Machine Learning for AI?",
  \url{https://aaai.org/Conferences/AAAI-18/invited-speakers/}},
current means to assess the quality of NNs focus on accuracy only,
encouraging behaviors that are good on average. This leaves out
a number of important properties and creates vulnerabilities in NNs
which can be exploited, \eg, in cyber-attacks to AI systems. On the
other side, Automated Reasoning has made, possibly less spectacular,
but steady progresses as well. Automated reasoning tools
provide staple techniques for hardware, software and protocol
verification in research as well as industrial applications ---
see, \eg,~\cite{harrison2009handbook}.
Researchers expanded the reach beyond the traditional ``comfort zone''
of discrete finite-state models to deal with cyber-physical systems
incorporating continuous and stochastic dynamics --- see,
\eg,~\cite{clarke2011statistical,kumar2012hybrid}. On top of this, the
community has been actively providing interesting
contributions towards solving the problem of NN verification as described in
Section~\ref{sec:stateofart}. Therefore, we believe that it is worth to
merge the two streams of research, by now largely independent, in
order to harness the power of automated verification whenever learning
methods require (formal) assurances related to their performances.

The second important challenge is to mate precision with scalability.
As shown, \eg,
by~\cite{DBLP:journals/aicom/PulinaT12,scheibler2015towards}, the main
barrier to applying off-the-shelf automated reasoning tools
to analyze NNs is that such tools hardly scale to deal with current
state-of-the art models: DNNs like AlexNet or GoogleLeNet
feature millions of parameters, resulting in
prohibitively large search spaces for
automated reasoning algorithms. This is why researchers
either restrict the scope of application of general-purpose tools ---
see, \eg,~\cite{DBLP:conf/cav/HuangKWW17} --- or develop
special-purpose solvers ---
see, \eg,~\cite{DBLP:conf/cav/KatzBDJK17,DBLP:conf/atva/Ehlers17}.
Even if SAT and MIP techniques are known to scale very well on problems
of considerable size, researchers using them also report limitations
in dealing with networks of moderate size given current demands.
Noticeably, a lot of research in the ML community has been carried out
to reduce the size of ML models that are
known to be highly redundant in terms of the number of
parameters. Two examples of such techniques are
model reduction and knowledge distillation. Model reduction methods
take an original model and use, for example, quantization techniques
to reduce its size  while preserving accuracy. An
extreme case of such quantization is binarizing most parameters of the
network~\cite{CourbariauxBD15,HubaraCSEB16}.
Similarly, knowledge distillation algorithms start with a full network
and build a new smaller network without a significant accuracy
loss. Conversely, on the Automated Reasoning side,
there is a need of effective tools. For instance, large
classes of NNs, \eg, NNs with ReLU activation functions,
can be encoded using linear arithmetic constraints enhanced with
simple logical constraints --- an approach considered
by~\cite{DBLP:conf/atva/Ehlers17}; other examples of specialization
include the Reluplex algorithm that analyzes only NNs with ReLU activation
functions~\cite{DBLP:conf/cav/KatzBDJK17} and, as mentioned
in~\cite{DBLP:journals/corr/abs-1709-06662,DBLP:journals/corr/abs-1802-08795}, the usage of
pseudo-Boolean constraints to encode binarized NNs in order to
leverage pseudo-Boolean solvers rather than a fully fledged SMT
solvers. Additionally, verification tools  should exploit structural
properties of NNs, \eg, the layered graph structure of the network,
and the parameter sharing features of convolutional layers.  For example,
abstraction techniques or clever decompositions can be used to verify
only critical parts of the models --- similarly to
what~\cite{DBLP:conf/cav/HuangKWW17} proposed.


Overall, we believe that the synergy between Machine Learning
and Automated Reasoning communities can be very fruitful. Automated
verification of NNs could be the new driving force for
theoretical and practical advancements in Automated Reasoning and,
at the same time, ML could benefit from powerful verification
techniques to generate  proofs of correctness for NNs.
Such techniques could even be used  to explain models, making them
more amenable to application in regulated contexts. Hybrid solutions
could also be adopted, where verification could be embedded into the
learning process, and learning could support effective
verification. In the following, we briefly outline some of the
directions that can be explored to improve on the current state of the art.
\begin{description}
\item[Common standards and APIs.] Appropriate standards are to be
  developed such that learning tools can be easily and interchangeably
  connected to verification ones. APIs are required in cases where
  integration should be tight: for   instance
  in~\cite{DBLP:conf/cav/PulinaT10}, repairing the NN required
  communication between verification and
  learning engines to leverage counterexamples and improve on learning.
\item[Benchmark collections.] Right now, most contributions cited in
  Section~\ref{sec:stateofart} consider their own case studies
  wherefore they provide specific results --- with MNIST database
  images being an exception. The creation and the
  dissemination of a library of  benchmarks ---- respecting the
  standards defined above --- will enable communities to share challenges and
  measure progress.
\item[Hybrid reasoning methods.] It is unlikely that complex,
  implements trained on dedicated hardware like modern DNNs
  can be verified by general purpose automated reasoning tools and
  hardware. Developers of automated reasoning techniques must realize
  this, and push innovation towards tools that can deal with, \eg,
  non linearities, non-algebraic activation functions, and complex
  architectures, possibly harnessing the power of the same computing
  devices used to train NNs.
\item[Transparent learning models.] The tendency towards creating
  networks whose size and complexity makes them hardly explainable
  should probably be reconsidered, at least for networks whose safety
  or security must be guaranteed in all expected working
  conditions. Therefore, we need the ability to design and train
  networks whose effectiveness is uncompromised, but whose
  verification is feasible. This is a goal that the Machine Learning
  community must acknowledge and
  consider~\cite{DBLP:journals/corr/abs-1801-05950}.
\end{description}

{\small
\bibliographystyle{named}
\bibliography{ijcai18}
}

\end{document}